\documentclass[10pt,twocolumn,letterpaper]{article}

\usepackage{cvpr}              
\usepackage{caption}  
\usepackage{pifont}
\usepackage{tabularx}
\usepackage{booktabs}
\usepackage{cuted}  
\usepackage[ruled,vlined]{algorithm2e}
\usepackage{amsmath}
\usepackage{xcolor}
\definecolor{cvprblue}{rgb}{0.21,0.49,0.74}
\usepackage[pagebackref,breaklinks,colorlinks,allcolors=cvprblue]{hyperref}

\makeatletter
\let\oldmaketitle\maketitle
\renewcommand{\maketitle}{
    \oldmaketitle
    \begin{strip}
        \centering
        \vspace{-12mm}
        \includegraphics[width=1.0\textwidth]{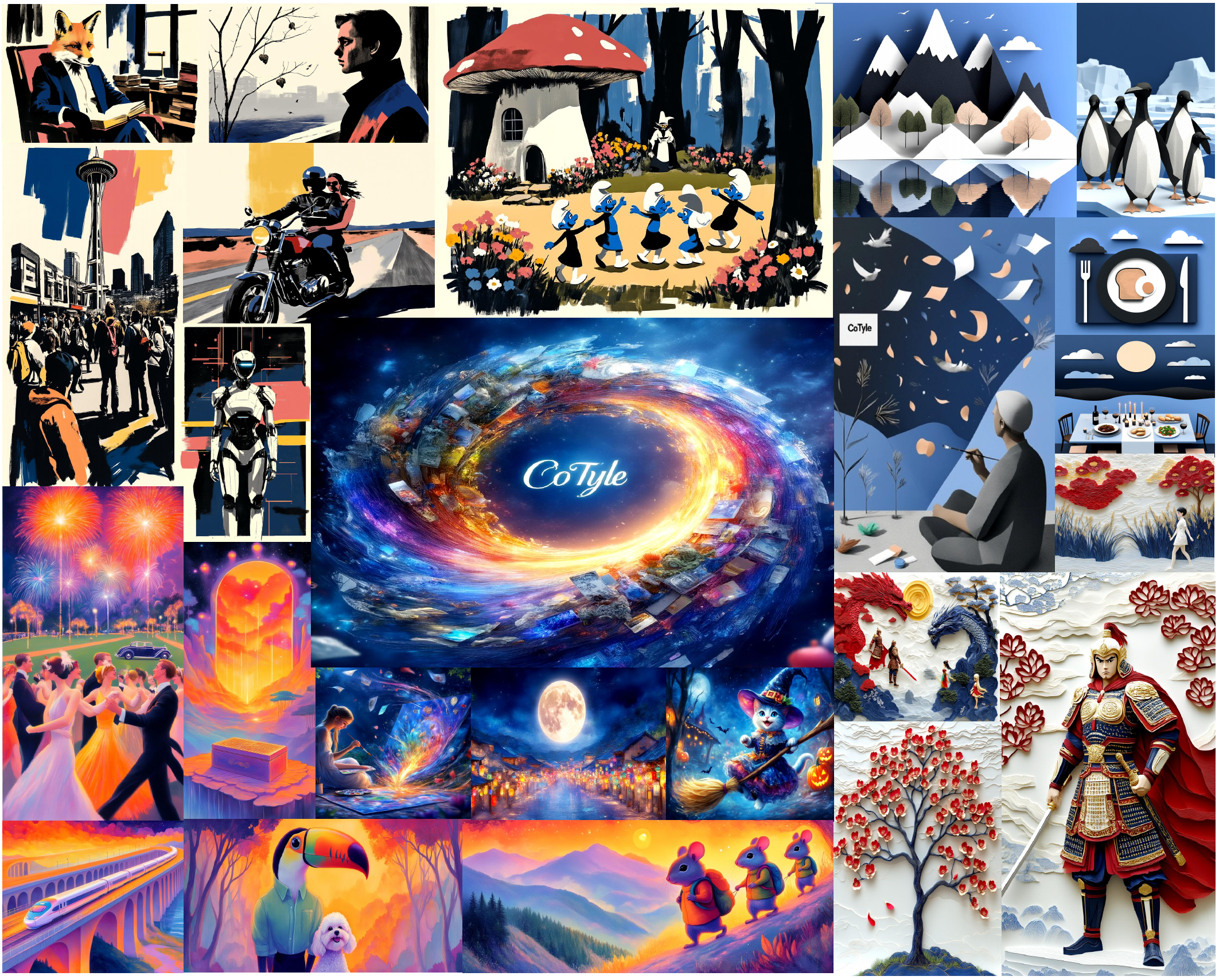}
        \vspace{-8mm}
        \captionof{figure}{\textbf{Visual demonstration of CoTyle.} The figure shows five groups of samples (top-left, top-right, bottom-left, bottom-right, center), each generated from a distinct style code, with consistent style within each group and distinct styles across groups.
        Our homepage can be found in \textit{\textcolor{deeppink}{https://kwai-kolors.github.io/CoTyle/}}.
        }
        \label{fig:teaser}
    \end{strip}
}
\makeatother

\definecolor{cvprblue}{rgb}{0.21,0.49,0.74}
\usepackage[pagebackref,breaklinks,colorlinks,allcolors=cvprblue]{hyperref}

\def\paperID{304} 
\def\confName{CVPR}
\def\confYear{2026}

\title{A Style is Worth One Code: Unlocking Code-to-Style Image Generation with Discrete Style Space}

\author{
Huijie Liu$^{1,2}$ \hspace{0.11em}
Shuhao Cui$^{2}$ \hspace{0.11em}
Haoxiang Cao$^{2,3}$ \hspace{0.11em}
Shuai Ma$^{1}$ \hspace{0.11em}
Kai Wu$^{2, \dag}$ \hspace{0.11em}
Guoliang Kang$^{1,\dag}$
\\
$^{1}$Beihang University \hspace{0.6em}
$^{2}$Kolors Team, Kuaishou Technology \hspace{0.6em}
$^{3}$South China Normal University \hspace{0.15em} \\
$^{\dag}$Co-Corresponding Author
}

\begin{document}

\definecolor{deeppink}{rgb}{1.0, 0.0, 0.5}

\maketitle


\let\thefootnote\relax
\footnotetext{This work was conducted during the author's internship at Kolors Team, Kuaishou Technology.}
\begin{abstract}
Innovative visual stylization is a cornerstone of artistic creation, yet generating novel and consistent visual styles remains a significant challenge. 
Existing generative approaches typically rely on lengthy textual prompts, reference images, or parameter-efficient fine-tuning to guide style-aware image generation, but often struggle with style consistency, limited creativity, and complex style representations.
In this paper, we consider the code-to-style image generation task, which aims to produce images with novel and consistent visual styles specified by only a numerical code.
To date, this field has only been primarily explored by the industry (e.g., Midjourney), with no open-source research from the academic community.
To fill this gap, we propose \textbf{CoTyle}, the first open-source method for this task.
Specifically, we first train a discrete style codebook from a collection of images to extract style embeddings.
These embeddings serve as conditions for a text-to-image diffusion model (T2I-DM) to generate stylistic images.
Subsequently, we train an autoregressive style generator on the discrete style embeddings to model their distribution, allowing the synthesis of novel style embeddings. 
During inference, a numerical style code is mapped to a unique style embedding by the style generator, and this embedding guides the T2I-DM to generate images in the corresponding style.
Extensive experiments validate that CoTyle effectively converts a numerical code into a style controller, demonstrating a style is worth one code.
Compared to existing methods, the stylized images generated by our method are more diverse and consistent, unlocking a vast space of reproducible styles from minimal input.
\end{abstract}    
\vspace{-1mm}
\section{Introduction}
\label{sec:intro}
Innovative visual stylization is a cornerstone of artistic creation, enabling the expression of unique identities and emotions across digital art, design, and media.
With generative models, \emph{e.g.,} diffusion models~\cite{flux2024, labs2025flux,rombach2022latentdiffusion,huang2024dialoggen,saharia2022photorealistic,podell2023sdxl}, users can generate high-fidelity images with arbitrary styles.

\begin{figure*}[tbp]
  \centering
  \vspace{-7mm}
  \begin{subfigure}{0.95\linewidth}
  \includegraphics[width=\linewidth]{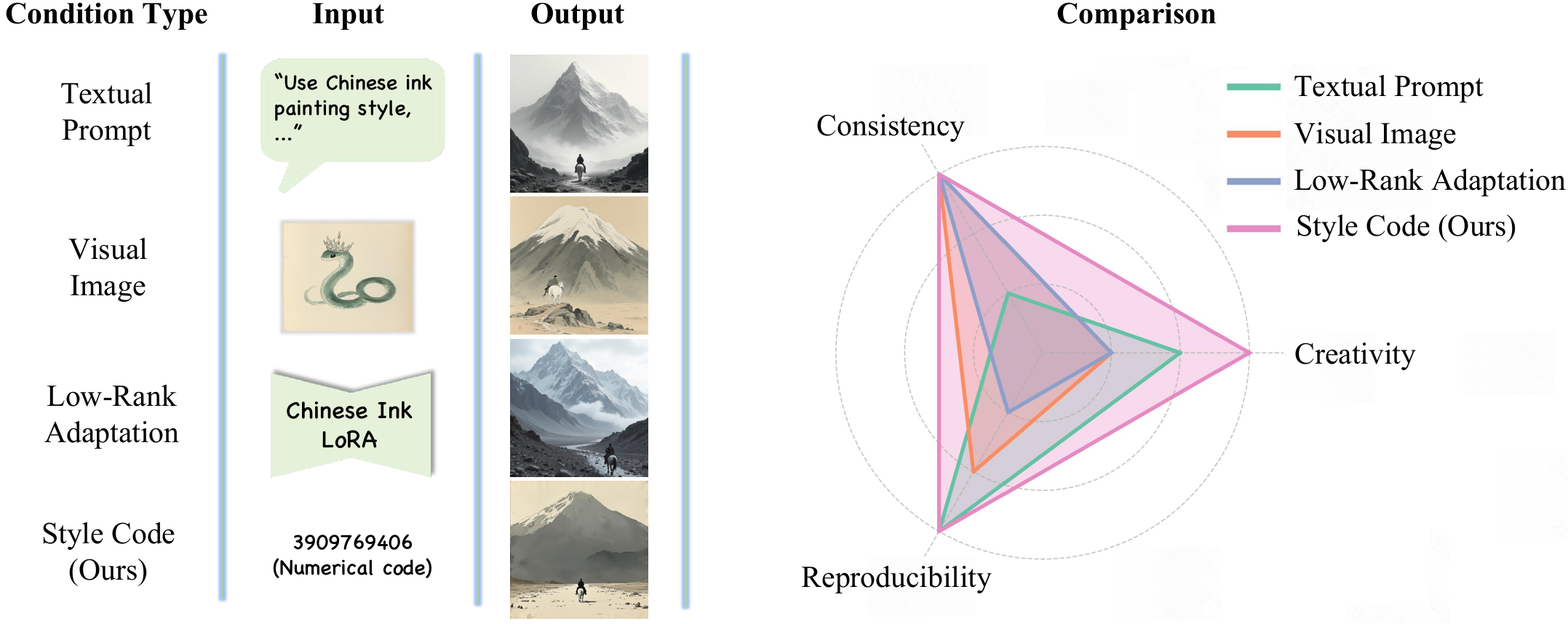} 
  \end{subfigure}
  \vspace{-1mm}
  \caption{
  Different to previous methods, CoTyle uses a numerical style code to represent a style, eliminating the need for complex prompts, images, or LoRAs, and allowing easy creation of unique styles just modifying the code. ``Creativity'', ``Consistency'', and ``Reproducibility'' refer to a model’s ability to (1) generate novel styles, (2) produce multiple images in the same style consistently, and (3) reproduce styles using simple, user-friendly style definitions.}
  \label{fig:motivation}
  \vspace{-1mm}
\end{figure*}

Existing works on stylistic image generation can be categorized into several groups according to the style specification type, including methods specifying styles by textural prompt, visual image, Low-Rank Adaptation (LoRA)~\cite{hu2022lora}, as illustrated in Fig.~\ref{fig:motivation}. The radar chart in the figure compares representative methods across three key evaluation dimensions, including \textit{consistency, creativity and reproducibility}. 
Methods specified style by textural prompts usually exhibit \textit{poor stylistic consistency}, as images generated from identical textual style descriptions often exhibit substantial variations in visual appearance. 
Other methods either leverage the style of reference image as the style definition ~\cite{wu2025uso,lei2025stylestudio,xing2024csgo,wang2024instantstyle-plus} or insert a pre-trained style LoRA~\cite{hu2022lora} into the generative model to enable specific style generation~\cite{frenkel2024blora,shah2024ziplora}.
Nevertheless, both methods rely on existing style images (either as reference images or LoRA training data), rendering them incapable of creating novel, unseen artistic styles, \emph{i.e.} they own \textit{poor creativity}.
Furthermore, transferring style information for reproducible generation via these methods requires sharing pixel-level reference images or complex LoRA weights, resulting in \textit{poor reproducibility}.
Thus, existing methods cannot simultaneously achieve high consistency, creativity and reproducibility.

To address these limitations, we propose code-to-style generation, a novel paradigm that uses a numerical style code to create novel styles.
This code also functions as both a unique identifier for style creation and consistent style reproduction.
To date, this task has only been primarily explored by industrial settings (e.g., Midjourney~\cite{mj}), with no open-source research from the academic community.
In response, we propose \textbf{CoTyle}, the first framework designed to realize code-to-style generation and democratize this capability for open-source community and academic research.

The core concept of CoTyle is to design a style generator capable of producing novel style embedding that guides the T2I-DM in generating images with a specific style. 
Specifically, we begin by training a discrete style codebook~\cite{esser2021vqgan, van2017vqvae, ma2025unitok} with contrastive loss on paired style images, with the objective of mapping images sharing the same style to similar distribution, while pushing apart the distributions of distinct styles.
This process allows the discrete codebook to pool stylistic information from image features while suppressing irrelevant content.
The trained codebook is integrated into a pre-trained T2I-DM. 
We train the T2I-DM to condition its generation on the codebook's output. 
This results in the T2I-DM capable of image-conditioned stylization.
To unlock code-to-style generation, we train an autoregressive transformer as a style generator. 
It learns the distribution of style indices derived from the style codebook under a next-token prediction objective, effectively modeling sequences of indices that represent coherent styles. 
Thus, the autoregressive model serves as an unconditional style generator, producing novel style index sequences.
During inference, a random token, which is fixed by input code, initializes the autoregressive generation process. 
The model predicts a complete sequence of indices. 
These indices act as conditioning signals to guide the T2I-DM, enabling the generation of images with a consistent, code-defined style.

To evaluate our method, we employ the CSD~\cite{somepalli2024csd} to assess both style consistency and diversity. Extensive experiments validate the efficiency of CoTyle, which generates images in target styles without requiring reference images, lengthy prompts, or model fine-tuning. 
Furthermore, we extend CoTyle to support image-driven stylized image generation and style interpolation.
As illustrated in Fig.~\ref{fig:teaser}, CoTyle addresses limitations existing in previous works: 1) similar to image-conditioned methods, it guarantees high stylistic consistency across multiple images generated with the same style code;
2) it enables the creation of novel artistic styles based on arbitrary style codes, achieving strong creativity; 3) the numerical style code serves as a concise and portable representation of the target style, ensuring excellent reproducibility for seamless sharing and reproducible generation.

In a nutshell, our contributions are as follows: 
\begin{itemize}
    \item We introduce code-to-style image generation, a novel task which enables the creation of diverse, consistent visual styles conditioned solely on a numerical style code.
    \item We propose \textbf{CoTyle}, the first open-source framework that achieves code-to-style generation by learning a discrete style codebook and an autoregressive style generator.
    Further, we expand it to facilitate image-conditioned generation and supports style interpolation.
    \item  We conducted extensive experiments that demonstrate the effectiveness of CoTyle. The results validate that a single code can serve as a powerful, compact style controller, unlocking a vast space of reproducible novel styles.
\end{itemize}

\section{Related Work}
\label{sec:related_work}
\textbf{Conditioned Image Generation.}
Conditioned image generation produces visual content conditioned on user-provided control signals.
Early methods~\cite{dalle3,saharia2022imagen,rombach2022latentdiffusion} established text-based conditioning for text-to-image generation, while subsequent work introduced richer signals.
ControlNet~\cite{zhang2023controlnet} and T2I-Adapter~\cite{mou2024t2i-adapter} provide structural guidance via edge/depth maps or sketches, and DreamBooth~\cite{ruiz2023dreambooth} with Textual Inversion~\cite{gal2022textualinversion} enable identity-specific generation from few-shot examples.
Common conditioning mechanisms include noise inversion~\cite{cao2025causalctrl, song2020ddim}, adapter modules~\cite{guo2024pulid,ye2023ipadapter,betterfit}, parameter-efficient fine-tuning~\cite{frenkel2024blora,shah2024ziplora,liu2025separate}, and token concatenation~\cite{tan2025ominicontrol,tan2025ominicontrol2,liu2025omnidish,boow,zhang2025grouprelativeattentionguidance}.
Few studies leverage vision-language models to encode visual information as prompt embeddings; while Qwen-Image~\cite{wu2025qwenimage} does so, it is primarily designed for image editing.

\textbf{Stylistic Image Generation.}
Stylistic image generation aims to produce images with specific style based on user-provided style references and text descriptions. 
Certain approaches~\cite{lei2025stylestudio,xing2024csgo,qi2024deadiff,wu2025uso,labs2025flux,wang2024instantstyle,wang2024instantstyle-plus} employ style images as visual conditions, while others~\cite{frenkel2024blora,shah2024ziplora} train style LoRA~\cite{hu2022lora} modules to steer the generation toward target styles. 
However, these signals require complex pixel-level maps or additional model parameters for representation, and more importantly, they lack the capability to create novel styles. In industrial applications, Midjourney~\cite{mj} addressed this limitation by introducing style code-based image generation, yet the absence of technical reports has hindered its adoption and development within the open-source community.

\begin{figure*}[htbp]
  \centering
  \vspace{-7mm}
  \begin{subfigure}{0.93\linewidth}
  \includegraphics[width=\linewidth]{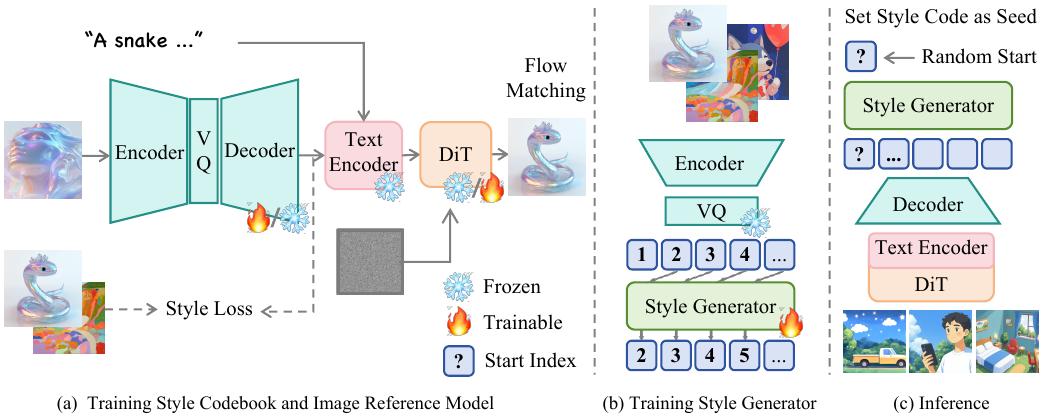}    
  \end{subfigure}
  \vspace{-1mm}
  \caption{\textbf{Overview of CoTyle}. (a) We first train a style codebook and an image generation model conditioned on style images. (b) Then, we use the corresponding codebook indices of the style images to train an autoregressive style generator. (c) During inference, a style code is used to randomly sample the first index and autoregressively predict the rest.}
  \label{fig:overview}
  \vspace{-1mm}
\end{figure*}

\section{Method}
\label{sec:method}
In this section, we propose \textbf{CoTyle}, the first open-source method for code-to-style generation, as shown in Fig.~\ref{fig:overview}.
CoTyle comprises three main components.
We begin by training a discrete style codebook with pairs of style images (Sec.~\ref{sec:method1}). 
The codebook can extract discrete style embeddings from reference images.
Using these style embeddings, we train a T2I-DM capable of generating images that share the same style as the reference image. (Sec.~\ref{sec:method2}). 
Finally, we train an autoregressive style generator to generate style indices, unlocking seed-to-style generation (Sec.~\ref{sec:method3}).

\subsection{Style Codebook}
\label{sec:method1}
Our core idea is to design an autoregressive style generator that produces style indices to guide the T2I-DM in generating images with a specific style. To realize this generator, we first train a discrete style codebook to act as a style extractor.
The use of a discrete codebook offers two main advantages: (1) the discrete indices align naturally with the next-token prediction objective of autoregressive modeling, and (2) the quantization process inherently suppresses irrelevant content information, which facilitates more effective pooling of style features from images.

However, unlike traditional codebooks~\cite{van2017vqvae,esser2021vqgan,ma2025unitok} used for image reconstruction, the style codebook is not designed to reconstruct the original image with high fidelity using discrete embedding. 
Instead, its purpose is to encode images with identical styles but different content into the same distribution, while encoding images with divergent styles into different distributions.
Thus, we employ a contrastive loss to train the model. 
Without the contrastive loss, the model would simply learn to map diverse styles to an identical embedding.
The codebook is trained using features extracted from a vision transformer (ViT)~\cite{dosovitskiy2020vit} and the loss function can be formulated as follows,
\begin{equation}
\resizebox{0.9\linewidth}{!}{%
    $\displaystyle
    \mathcal{L}_{\text{contrast}} = \frac{1}{B} \sum_{i=1}^{B} \left[
        y_i \cdot (1 - s_i)^2 + 
        (1 - y_i) \cdot \left( \text{ReLU}(s_i - m) \right)^2
    \right],
    $%
}
\end{equation}

\begin{equation}
s_i = \frac{\mathcal{F}(\mathbf{v}_{1,i}) \cdot \mathbf{v}_{2,i}}{\|\mathcal{F}(\mathbf{v}_{1,i})\| \|\mathbf{v}_{2,i}\|},
\end{equation}
where $B$ is the batch size,
$y_i \in \{0, 1\}$ is the label for the $i$-th sample pair (1 for identical style and 0 for different style),
 $m$ is the margin parameter that defines the minimum desired separation for negative pairs,
$\mathbf{v}_{1,i}$ and $\mathbf{v}_{2,i}$ are the ViT features of the two samples in the $i$-th pair, and $\mathcal{F}(\cdot)$ denotes the style codebook.

Further, we find that adding a reconstruction loss is essential to avoid codebook collapse during training (Sec.~\ref{sec:exp}). 
This is because we need to leverage the capabilities of the pre-trained vision language model (Sec.~\ref{sec:method2}), and our style embeddings should remain close to the image embeddings output by the VLM's image encoder.
\begin{equation}
\mathcal{L}_{\text{recon}} = \frac{1}{N} \sum_{i=1}^{N} \left[\frac{\mathcal{F}(\mathbf{v}_{1,i}) \cdot \mathbf{v}_{1,i}}{(\|\mathcal{F}(\mathbf{v}_{1,i})\| \|\mathbf{v}_{1,i}\|}\right]^2,
\end{equation}

Similar to traditional vector-quantized methods~\cite{van2017vqvae, esser2021vqgan, ma2025unitok}, we employ a commit loss and a codebook loss. 
The vector quantization loss $\mathcal{L}_{\text{vq}}$ is defined as the sum of them. 
The final overall loss function is formulated as,
\begin{equation}
\mathcal{L}_{\text{style}} = \mathcal{L}_{\text{contrast}} + \alpha \mathcal{L}_{\text{recon}} + \beta \mathcal{L}_{\text{vq}},
\end{equation}
where $\alpha$ and $\beta$ are weighting coefficients.

\subsection{T2I-DM Conditioned on Style Codebook}
\label{sec:method2}
To utilize the embedding quantized by the style codebook as style condition, we integrate the codebook into a T2I-DM as shown in Fig.~\ref{fig:overview} (a).
Unlike traditional style transfer methods~\cite{gao2024styleshot,wu2025uso,ye2023ipadapter,qi2024deadiff}, we argue that style information should not be narrowly defined as color, but encompasses rich semantic features.
Thus, we treat style embeddings as a form of textual input and injecting them into the Diffusion Transformer (DiT~\cite{peebles2023dit}) through textual branch. 
Specifically, we employ a vision language model (VLM)~\cite{qwen2.5-VL} as our text encoder, while the style embedding replaces the original image features.
This method can help the T2I-DM learn style information that better aligns with human perception.

During training, for each pair of images $x_1$ and $x_2$ sharing the same style, we extract the ViT features $\mathbf{v}_1$ from $x_1$ and quantize it into style embedding $\mathcal{F}(\mathbf{v}_1)$. 
We train the T2I-DM~\cite{peebles2023dit} conditioned on both the style embedding $\mathcal{F}(\mathbf{v}_1)$ and the text prompt $y_2$ (which corresponds to $x_2$) to generate the target image $x_2$ by rectified flow matching~\cite{lipman2022flowmatching}.
After training, DiT can generate images with specific styles based on the style embeddings output by the codebook.

Notably, although CoTyle is designed for code-to-style generation, it inherently supports image-conditioned generation and outperforms existing methods (Sec.~\ref{sec:exp}).

\subsection{Code-to-Style Image Generation}
\label{sec:method3}
We now construct a T2I-DM conditioned on embeddings from a style codebook.
However, these embeddings are derived from existing images, thus limiting the creation of novel styles.
To enable code-to-style generation, we need to train an unconditional style generator to create novel styles.  
As shown in Fig.~\ref{fig:overview}~(b), for a given image, we extract its ViT feature $v$ and acquire the corresponding discrete indices from the style codebook. 
These indices are then used to train an autoregressive model via next token prediction~\cite{shannon1949ntp}, effectively learning the distribution of style features. 

\begin{figure*}[htbp]
  \centering
  \vspace{-7mm}
  \begin{subfigure}{0.97\linewidth}
  \includegraphics[width=\linewidth]{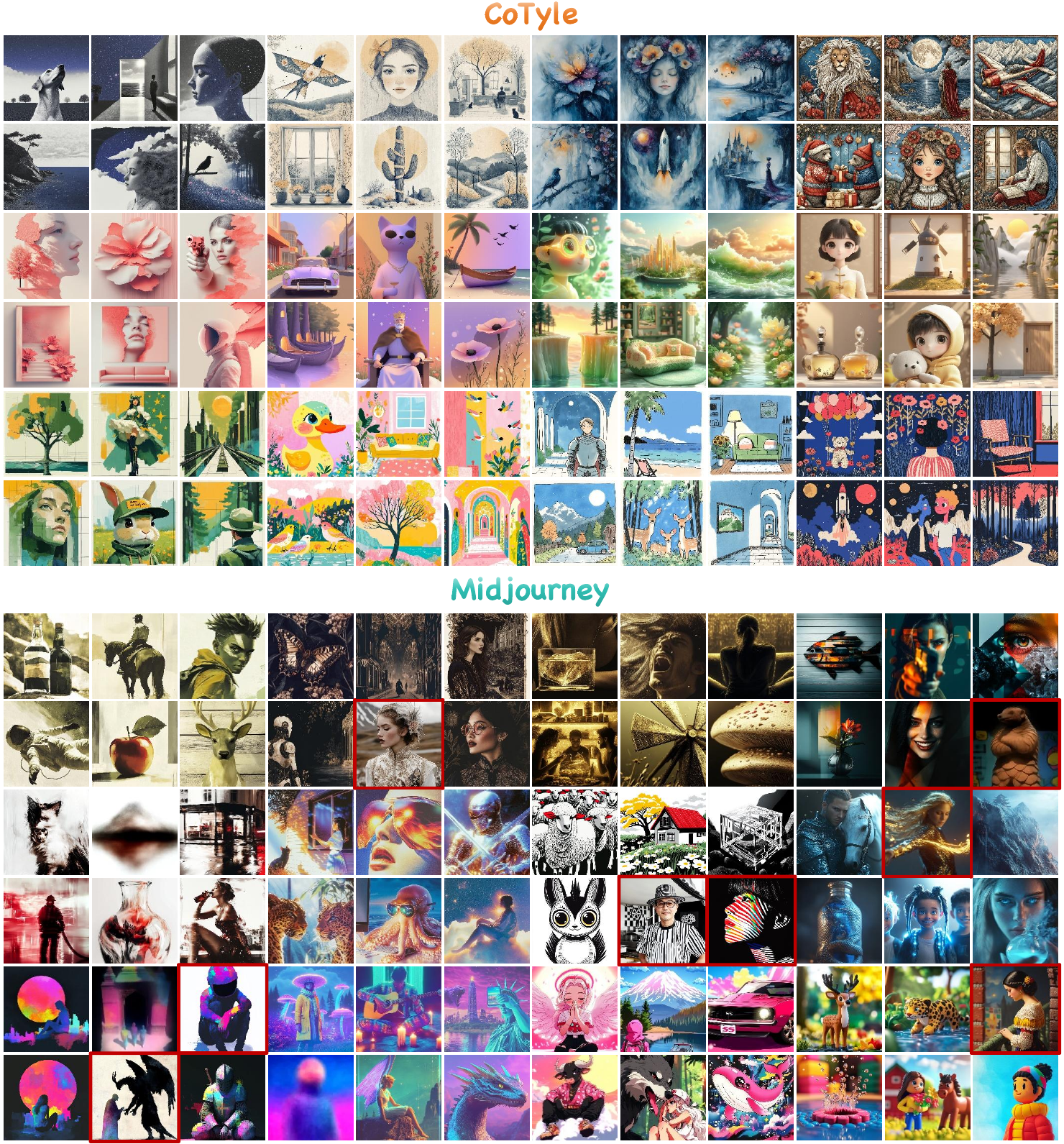}  
  \end{subfigure}
  \vspace{-1mm}
  \caption{Qualitative comparison with Midjourney~\cite{mj} on code-to-style generation. Each image set (2×3 grid) is generated from the same style code. Red boxes highlight cases with suboptimal style consistency.}
  \label{fig:ablation_mj}
\end{figure*}
Fig.~\ref{fig:overview} illustrates the inference process.
A user-provided numerical code initializes the random seed. 
Using this fixed seed, the model samples an initial token $I_0$ and autoregressively generates the subsequent $N-1$ tokens, where $N$ is a hyperparameter defining the total token sequence length. 
Using these indices, we retrieve the corresponding vectors from the codebook and decode them into style embeddings. 
Subsequently, we generate images with specific styles in a manner similar to the approach described in Sec.~\ref{sec:method2}.
The pseudocode for the inference process is presented in Alg.~\ref{alg:inference}.

To further enhance the intensity and diversity of the style, we propose a sampling strategy.
We analyzed the frequency of codebook index across a collection of images, and find that certain indices exhibit significantly higher selection frequencies than others (see Appendix for details). 
Empirical results (Fig.~\ref{fig:ablation_highfreq}) show that high-frequency indices represent a form of ``arbitrary'' style information, just like meaningless placeholders. 
Sampling exclusively from high-frequency indices yields images with no specific style. 
Therefore, during inference, we down-weight the logits corresponding to these indices by multiplying a suppression coefficient.  
The suppression coefficient can be formulated as,

\begin{algorithm}[tbp]
\caption{Code-to-Style Inference}
\label{alg:inference}
\textbf{Inputs:} style code $n$, prompt $y$ \\
\textbf{Given:} style codebook $C$, codebook size $K$, number of tokens $N$, suppression coefficient $s(i)$ \\
$x_t \sim \mathcal{N}\{0,1\}$ \\
SetSeed($ n $) \\
$I_{0} \sim \mathcal{U}\{0, \ldots, K\}$ \\

\For{$i = 1 \ldots N$}{
    $I_i$ = AR$(I_{0 : i-1}, s(i))$\;
}
c = Lookup($C, I_{0 \ldots N}$) \\

$x_0$ = RectifiedFlow($x_t, [c, y]$) \\

\textbf{Return:} $x_0$
\end{algorithm}

\begin{equation}
s(i) =
\begin{cases} 
1, & \text{if } f(i) < \tau \\
e^{-k(f(i)-\tau)}, & \text{if } f(i) \geq \tau
\end{cases}
\end{equation}
where $i$ is the index and $f(\cdot)$ denotes mapping index to frequency value. $\tau$ is a threshold and $k$ is a hyperparameter.

\section{Experiment}
\label{sec:exp}
\subsection{Implementation Details}
During the codebook training phase, we set the codebook vocabulary size to 1,024 and the embedding dimension to 64. The model was trained for 20,000 steps with a batch size of 128 and a learning rate of 1e-5.
In the DiT training stage, we initialized the model from pretrained Qwen-Image~\cite{wu2025qwenimage} and trained it for 60,000 steps with a batch size of 64 and a learning rate of 4e-6.
For the style generator, we employed the Qwen2-0.5B~\cite{qwen2} architecture but trained it from scratch without loading any pre-trained weights. This component was trained for 100,000 steps with a batch size of 64 and a learning rate of 1e-5.
We resized style references to 392×392 during training, as stylistic features are robust to geometric transformations.
Each image was then encoded into 196 style tokens, represented by 196 discrete style indices (\textit{i.e.} $N=196$).

\begin{table*}[htbp]
\centering
\vspace{-7mm}
\caption{
\textbf{
Quantitative comparison with other methods.}
``~\textsuperscript{*}~'' denotes CoTyle conditioned on reference images. Bold highlights the best score; underlines ``\underline{~~~~}'' indicate the second-highest, omitted in code-to-style evaluation for brevity.
CLIP-T measures text-image alignment.
For image-conditioned methods, diversity is constrained by the reference image and is neither measurable nor meaningful to evaluate.
}
\small
\begin{tabularx}{0.94\textwidth}{@{} 
    >{\raggedright\arraybackslash}m{2.8cm}
    >{\centering\arraybackslash}p{1.83cm}
    >{\centering\arraybackslash}p{1.83cm}
    >{\centering\arraybackslash}p{1.83cm}
    >{\centering\arraybackslash}p{1.83cm}
    >{\centering\arraybackslash}p{1.83cm}
    >{\centering\arraybackslash}p{1.9cm}@{}}
    \toprule
    Methods & Open-Source & Condition & Diversity~$\uparrow$ & Aesthetics~$\uparrow$ & CLIP-T~$\uparrow$ & Consistency $\uparrow$ \\
    \midrule
    StyleStudio~\cite{lei2025stylestudio} & \ding{51} & Image & - & 0.5074 & 0.3168 & 0.4711 \\
    CSGO~\cite{xing2024csgo} & \ding{51} & Image & - & 0.6283 & 0.3017 & 0.5540 \\
    USO~\cite{wu2025uso} & \ding{51} & Image & - & 0.7153 & \textbf{0.3331} & 0.4395 \\
    Flux-Kontext~\cite{labs2025flux} & \ding{51} & Image & - & \textbf{0.7636} & 0.3056 & 0.4222 \\
    InstantStyleXL~\cite{wang2024instantstyle-plus} & \ding{51} & Image & - & 0.7135 & 0.3134 & \underline{0.5753} \\
    \textbf{CoTyle\textsuperscript{*} (Ours)} & \ding{51} & Image & - & \underline{0.7178} & \underline{0.3230} & \textbf{0.5791} \\
    \midrule
    Midjourney~\cite{mj} & \ding{55} & Code & \textbf{0.8088} & 0.5948 & 0.3090 & 0.4734 \\
    \textbf{CoTyle (Ours)} & \ding{51} & Code & 0.7764 & \textbf{0.7173} & \textbf{0.3119} & \textbf{0.6007} \\
    \bottomrule
\end{tabularx}
\label{tab:ref}
\end{table*}

\subsection{Benchmark} 
For code-to-style generation, we randomly sampled 500 codes, generating 4 images per code, resulting in 2,000 images for evaluation.
For image-conditioned generation, we constructed an evaluation set of 500 prompt and reference image pairs. 

Following~\cite{wu2025uso, xing2024csgo}, we adopted CSD~\cite{somepalli2024csd} to evaluate image-conditioned methods.
For style consistency (Consistency), we computed the CSD score among images generated from the same code for CoTyle and Midjourney~\cite{mj}.
For image-conditioned methods~\cite{wu2025uso, wang2024instantstyle-plus, xing2024csgo, lei2025stylestudio, qi2024deadiff}, we measured the CSD score between their outputs and reference images.
To evaluate style diversity (Diversity), we randomly sampled multiple codes, generated corresponding images, and computed pairwise CSD scores among them.
We define the Diversity score as 1 minus the CSD score, so that higher values indicate greater diversity.
The Diversity score measures the degree of stylistic diversity across styles corresponding to different codes. 
Therefore, this metric is specifically designed for the code-to-style task.
Additionally, we evaluate text-image alignment using CLIP text-image similarity (CLIP-T)~\cite{wu2025uso, xing2024csgo, lei2025stylestudio} and assess the aesthetic quality of generated images via the recently popular QualityCLIP (Aesthetics)~\cite{agnolucci2024quality}.


\subsection{Baselines}
For the code-to-style task, it is currently the only approach comparable to Midjourney~\cite{mj}. 
Since Midjourney is not open-source, we manually collected 500 images from the Midjourney website for evaluation. 
Given that CoTyle also supports style generation conditioned on reference images, we have compared it against several popular style reference methods, including ~\cite{lei2025stylestudio, xing2024csgo, wu2025uso, flux2024,wang2024instantstyle-plus}.
\subsection{Results}

\textbf{Code-to-style image generation.}
We presented a comparison between CoTyle and the closed-source Midjourney~\cite{mj} on the code-to-style task in Fig.~\ref{fig:ablation_mj} and Tab.~\ref{tab:ref}, demonstrating that our method achieves significantly superior style consistency.
CoTyle achieved better style consistency, where different images generated from the same code exhibited highly similar styles.
However, CoTyle showed slightly inferior diversity, which may be attributed to the insufficient breadth of the training dataset, constraining CoTyle's ability to learn a wider spectrum of stylistic variations.
This represented a potential direction for future improvement.

\begin{figure*}[tbp]
  \centering
  \vspace{-3mm}
  \begin{subfigure}{0.97\linewidth}
  \includegraphics[width=\linewidth]{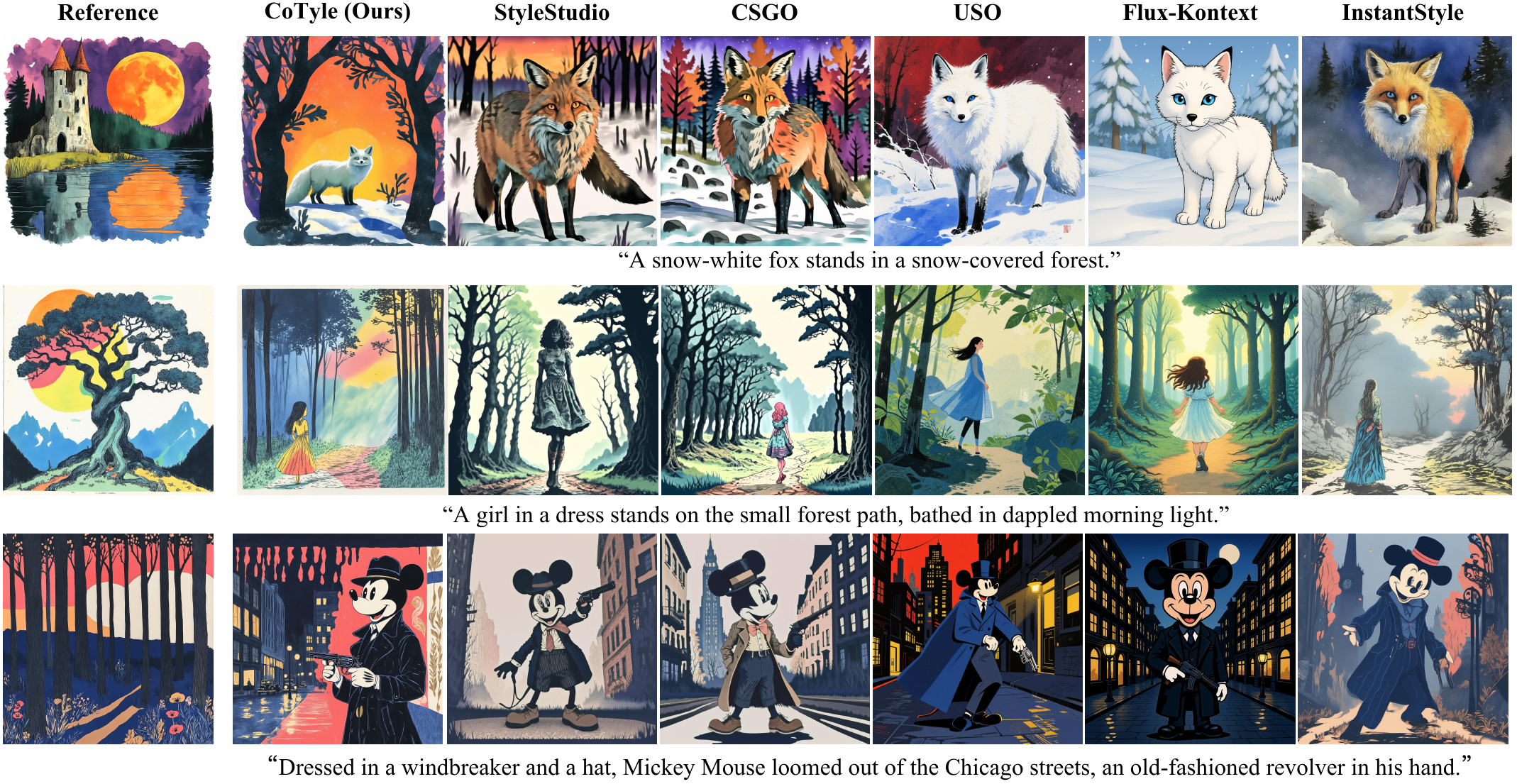}    
  \end{subfigure}
  \vspace{-1mm}
  \caption{Qualitative comparison. 
  CoTyle is not only capable of generation conditioned on style codes but also supports style images. Our model can faithfully follow the input text while simultaneously generating the specified style.
  }
  \label{fig:compare}
\end{figure*}

\textbf{Comparison with image-conditioned methods.}
We compared our method with existing image-conditioned approaches.  
CoTyle demonstrated significant advantages in style consistency.  
For instance, in Fig.~\ref{fig:compare}, the images generated by our method exhibited more accurate color tones.  
Furthermore, in the example shown in Row 2, where the image edges featured milky-white borders, our method successfully learned this stylistic representation, while other methods showed no responsiveness to this style.

\begin{figure}[tbp]
  \centering
  \vspace{-2mm}
  \begin{subfigure}{0.97\linewidth}
  \includegraphics[width=\linewidth]{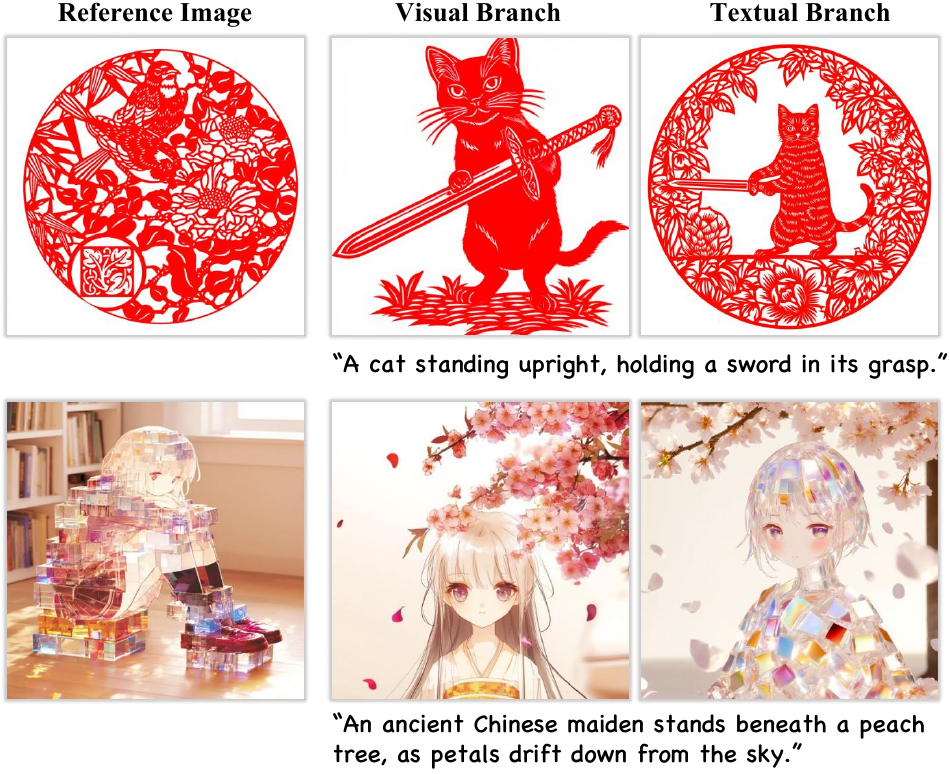}    
  \end{subfigure}
  \vspace{-1mm}
  \caption{
  We compare injecting style through textual branch with the existing method through visual branch. 
  Injecting style from the textual branch better preserves semantic information.}
  \vspace{-2mm}
  \label{fig:branch}
\end{figure}
\subsection{Ablation}
\vspace{-1mm}
\textbf{Effect of style injection through textual branch.}
CoTyle leverages a VLM as the text encoder to inject style information through the textual branch.
For comparison, we implement style injection through the visual branch using the OminiControl~\cite{tan2025ominicontrol, tan2025ominicontrol2}, \textit{i.e.}, by concatenating noise features with style condition features along the token dimension.
The quantitative results in Tab.~\ref{tab:ablation_text} confirm this perspective. 
As shown in the first row of Fig.~\ref{fig:branch}, visual branch injection captures the red tones of the paper-cut style but fails to recognize semantic-style elements (e.g., circular contours). 
For the second row, visual branch injection only captures warm-toned cues, whereas textual branch injection enables the model to correctly generate a human figure composed of crystal blocks.

\begin{table}[tbp]
    \centering
    \vspace{-1mm}
    \caption{Comparison of injecting style condition to DiT through visual branch and textual branch. }
    \small
    \begin{tabularx}{0.47\textwidth}{@{} 
      >{\raggedright\arraybackslash}m{2.0cm}
      >{\centering\arraybackslash}m{1.6cm}
      >{\centering\arraybackslash}m{1.6cm}
      >{\centering\arraybackslash}m{1.6cm}@{}}
      \toprule
      Methods & Aesthetics~$\uparrow$ & CLIP-T~$\uparrow$ & Consistency~$\uparrow$\\ 
      \midrule
      Visual Branch & 0.7175 & \textbf{0.3255} & 0.5306\\
      Textual Branch & \textbf{0.7178} & 0.3230 & \textbf{0.5791} \\
      \bottomrule
    \end{tabularx}
    \vspace{-1mm}
    \label{tab:ablation_text}
\end{table}
\begin{table}[tbp]
    \centering
    \small
    \vspace{-2mm}
    \caption{Effect of style loss $\mathcal{L}_{\text{style}}$. ``w/o. negative samples'' indicates that each data pair  sharing an identical style.}
    \begin{tabularx}{0.49\textwidth}{@{} 
      >{\raggedright\arraybackslash}m{2.93cm}
      >{\centering\arraybackslash}m{1.45cm}
      >{\centering\arraybackslash}m{1.22cm}
      >{\centering\arraybackslash}m{1.0cm}@{}}
      \toprule
      Loss & Aesthetics~$\uparrow$ & CLIP-T~$\uparrow$ & Consistency~$\uparrow$\\  
      \midrule
      $\mathcal{L}_{\text{style}}$ & \textbf{0.7178} & 0.3230 & \textbf{0.5791}\\
      w/o. negtive samples & 0.7174 & \textbf{0.3260} & 0.4890 \\
      w/o. $L_{recon}$ & 0.7001 & 0.3237 & 0.4102\\
      \bottomrule
    \end{tabularx}
    \vspace{-1mm}
    \label{tab:ablation_loss}
\end{table}
\begin{table}[tbp]
    \centering
    \footnotesize
    \vspace{-1mm}
    \caption{Effect of high-frequency suppression~$s(i)$.}
    \vspace{-1mm}
    \begin{tabularx}{0.494\textwidth}{@{} 
      >{\raggedright\arraybackslash}m{1.08cm}
      >{\centering\arraybackslash}m{1.45cm}
      >{\centering\arraybackslash}m{1.4cm}
      >{\centering\arraybackslash}m{1.40cm}
      >{\centering\arraybackslash}m{1.10cm}@{}}
      \toprule
      Methods & Diversity~$\uparrow$ & Aesthetics $\uparrow$& CLIP-T $\uparrow$ & Consistency~$\uparrow$ \\ 
      \midrule
      CoTyle & \textbf{0.7764} & 0.7173 & 0.3119 & \textbf{0.6007} \\
        w/o. $s(i)$& 0.7488 & \textbf{0.7177} & \textbf{0.3210} & 0.5301 \\
      \bottomrule
    \end{tabularx}
    \label{tab:ablation_si}
    \vspace{-1mm}
\end{table}

\begin{figure}[tbp]
  \centering
  \begin{subfigure}{0.97\linewidth}
  \includegraphics[width=\linewidth]{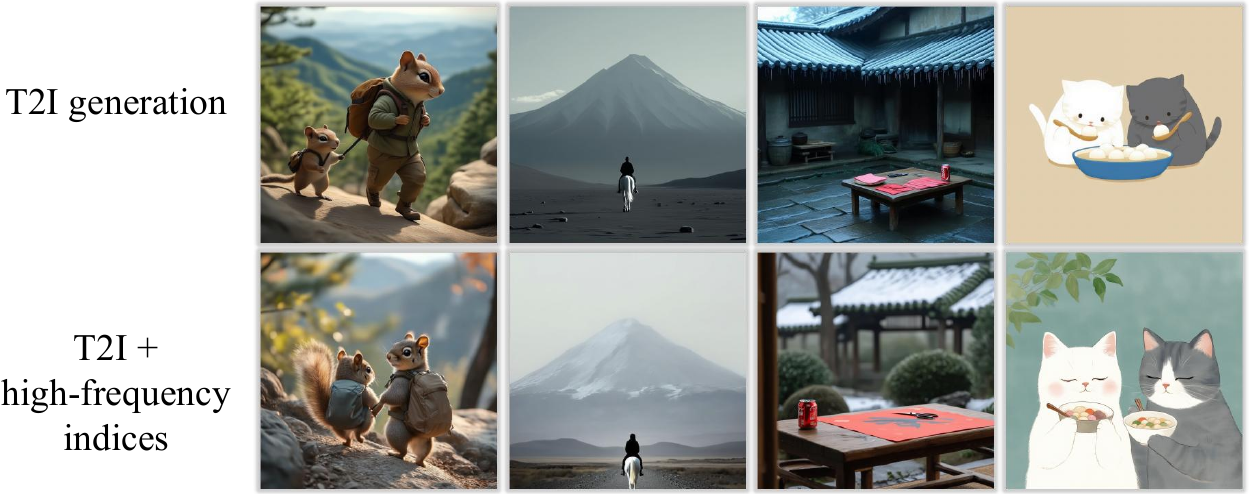}    
  \end{subfigure}
  \caption{
  Sampling solely from high-frequency indices yields style-less images. Row 1 shows the results of vanilla T2I-DM without any style indices, and Row 2, guided by high-frequency indices, produces results nearly identical to T2I-DM.}
  \label{fig:ablation_highfreq}
\end{figure}
\textbf{Effect of style loss.}
During style codebook training, the style loss $L_{style}$ comprised a contrastive loss for style learning and a reconstruction loss.  
We validated the necessity of both the contrastive and reconstruction components, 
The contrastive loss helped the model better extract style information, enhancing style consistency, while the reconstruction loss aligns the style embedding with the original ViT feature distribution, improving overall performance.
The results in Tab.~\ref{tab:ablation_loss}  support this hypothesis.

\textbf{Effect of frequency suppression.}
We found that sampling exclusively from high-frequency indices resulted in images devoid of stylistic diversity, as shown in Fig.~\ref{fig:ablation_highfreq}.  
We hypothesized that high-frequency indices might function as meaningless placeholders.
Therefore, we propose high-frequency suppression strategy. Tab.~\ref{tab:ablation_si} illustrates that without this strategy, the generated images exhibit low style diversity, as most style codes tend to produce photorealistic results. More analyses are provided in the Appendix.

\subsection{Style Interpolation}
\begin{figure}[tbp]
  \centering
  \begin{subfigure}{0.97\linewidth}
  \includegraphics[width=\linewidth]{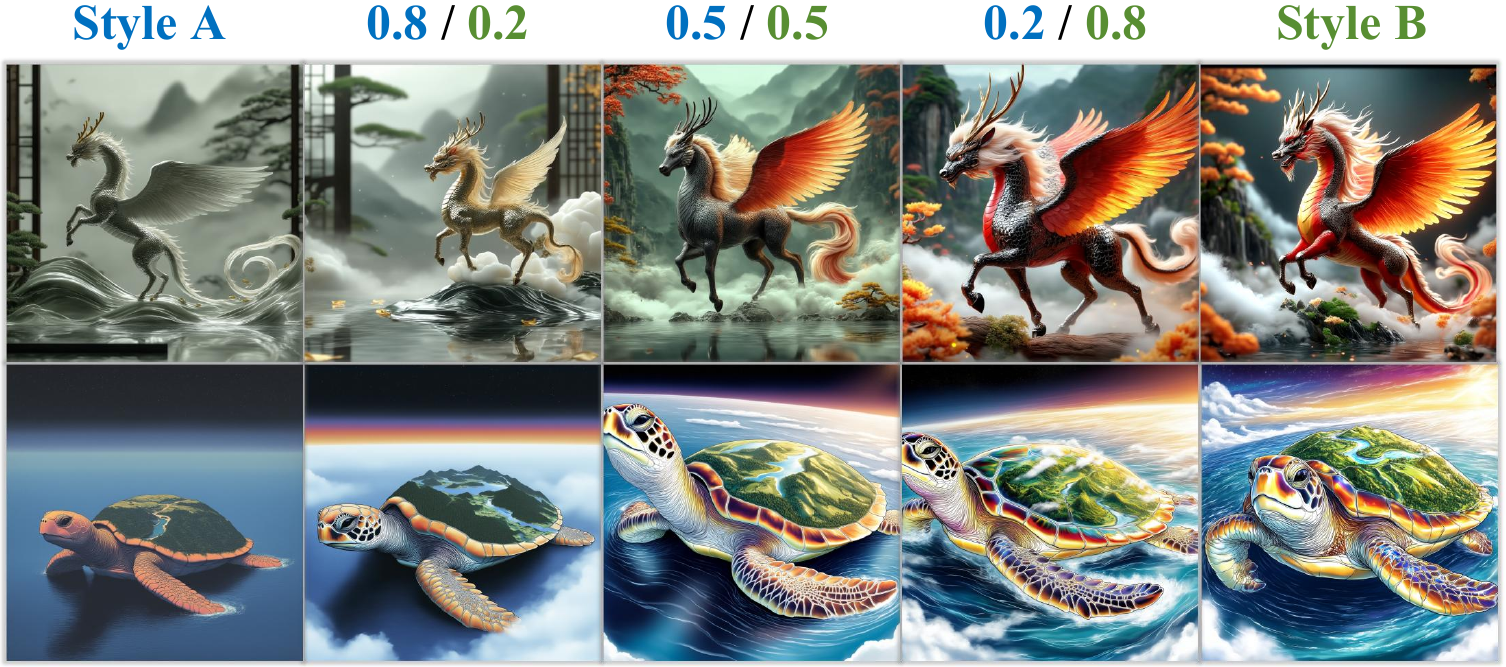} 
  \end{subfigure}
  \caption{
  Style interpolation. The leftmost and rightmost images represent two distinct styles. CoTyle enables smooth style interpolation by linearly combining multiple style indices according to user-specified weights. See more results in Appendix.}
  \label{fig:merge}
\end{figure}

Within CoTyle, each style is composed of $N$ indices predicted by a style generator.
Thus, fusion of different styles can be achieved by combining subsets of indices from each.
The intensity of the constituent styles is conditioned by adjusting their respective sampling ratios.
This combinatorial approach supports the blending of both predefined style codes and styles extracted from user-provided pixel images.
As illustrated in Fig.~\ref{fig:merge}, the leftmost and rightmost images represented two distinct styles.
By varying the proportion of their indices in the combination, different degrees of stylistic interpolation between them were achieved. 
This result reveals the composability of style features, suggesting that CoTyle may provide insights for controlling image styles.

\vspace{-2mm}
\section{Conclusion}
In this paper, we introduce code-to-style image generation, a novel task that enables stylistic image generation through numerical style codes. 
We propose CoTyle, the first open-source framework to support this task. 
CoTyle employs a dedicated style generator trained to produce novel style embedding from numerical style codes. 
Then, a diffusion model generates images with specific style conditioned on the style embedding.
Further, we extend CoTyle to support additional functionalities, including image-conditioned style generation and style interpolation.
Extensive experiments demonstrate that CoTyle effectively maps discrete codes to diverse visual styles. 
Our work establishes opens up future research in discrete stylistic representations.

{
    \small
    \bibliographystyle{ieeenat_fullname}
    \bibliography{main}
}
\clearpage
\setcounter{page}{1}
\maketitlesupplementary

\setcounter{figure}{0}
\setcounter{table}{0}
\setcounter{equation}{0}

\renewcommand{\thefigure}{A\arabic{figure}} 
\renewcommand{\thetable}{A\arabic{table}}   
\renewcommand{\theequation}{A\arabic{equation}} 

\section{Experimental Details}
In this chapter, we introduce the settings of the baselines we compared against.

When validating style-code-conditioned image generation, we compared our method with MidJourney~\cite{mj}. Since MidJourney's code is not open-sourced, we manually collected all test data from the MidJourney website.

When validating image generation conditioned on style images, we compared our method with state-of-the-art style image-based approaches, including StyleStudio~\cite{lei2025stylestudio}, CSGO~\cite{xing2024csgo}, USO~\cite{wu2025uso}, Flux-Kontext~\cite{flux2024,labs2025flux}, and InstantStyle~\cite{wang2024instantstyle}. 
For StyleStudio~\cite{lei2025stylestudio}, we set the classifier-free guidance (CFG)~\cite{ho2022cfg} to 5.0 and the number of denoising steps to 50. 
For the style scale controlling style intensity, we used the default value of 1.0 provided by the official implementation. 
For CSGO~\cite{xing2024csgo}, we set CFG to 10.0 and the denoising steps to 50. 
During generation, since no image condition is required, we followed the official recommendation and set the ControlNet~\cite{zhang2023controlnet} conditioning scale to 0.01 and the style scale to 1.0. 
For USO~\cite{wu2025uso}, we set CFG to 4.0 and the denoising steps to 25. For Flux-Kontext~\cite{flux2024,labs2025flux}, we set CFG to 2.5 and, following the procedure in its paper, used the style image as condition while prefixing the prompt with ``Using this style, ...''. 
For InstantStyle~\cite{wang2024instantstyle-plus}, we set CFG to 5.0, the denoising steps to 30, and the style scale to 1.0. 
For all the above methods, we loaded the official pre-trained model weights.

\section{Analysis of High-frequency Index}
To further investigate the learned style codebook, we analyze the utilization frequency of its indices to uncover potential biases or emergent patterns in how the model leverages its discrete representation space. Specifically, we construct a diverse dataset of stylistically varied images and encode each image using the trained codebook to extract the sequence of discrete indices that best reconstruct its style. We then aggregate the occurrence counts of each codebook index across the entire dataset, yielding a distribution that reflects the relative usage of each latent token.

\begin{figure}[tbp]
  \centering
  \begin{subfigure}{0.97\linewidth}
  \includegraphics[width=\linewidth]{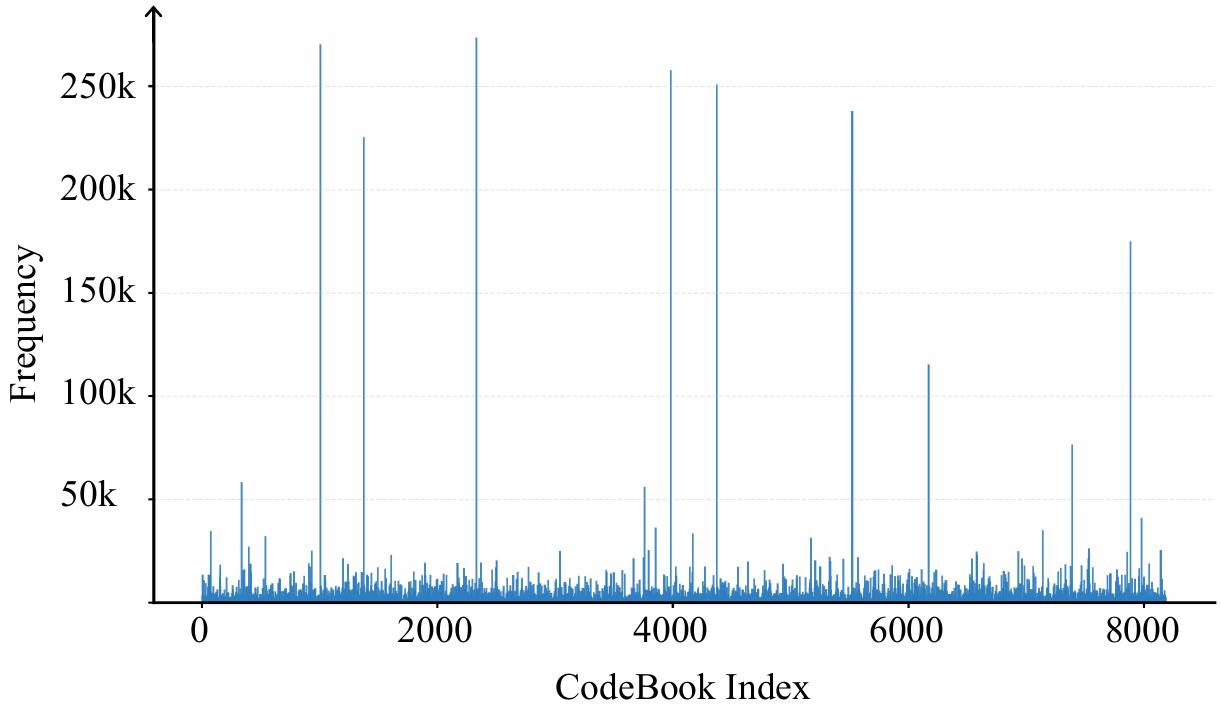}    
  \end{subfigure}
  \caption{
  \textbf{Frequency distribution of style codebook indices.} A batch of images is encoded using the style codebook, and the selection frequency of all indices is calculated.}
  \label{fig:freq}
\end{figure}

As illustrated in Fig.~\ref{fig:freq}, where the horizontal axis denotes the codebook index and the vertical axis represents the normalized selection frequency, we observe a pronounced long-tail distribution: a small subset of indices (approximately 5–10\% of the total) is selected with significantly higher frequency than the vast majority of others. This imbalance suggests that the model disproportionately relies on a limited number of tokens to encode stylistic information, rather than distributing semantic content evenly across the codebook.

Motivated by this observation, we conduct a detailed analysis in Sec.~\ref{sec:method3} of the main paper to determine whether these high-frequency indices correspond to interpretable stylistic attributes—such as brushstroke texture, color palette, or compositional layout. Through controlled ablations, we replace these dominant indices with random or shuffled values and observe minimal degradation in reconstruction quality or perceptual style fidelity. Furthermore, when we isolate the outputs generated using only the top-k high-frequency indices, the resulting images retain strong stylistic coherence despite the absence of low-frequency tokens. These findings strongly indicate that the high-frequency indices do not encode specific, semantically meaningful stylistic features. Instead, they function as placeholder tokens, essentially serving as generic, high-probability surrogates that the model defaults to for efficiency, likely due to overfitting or insufficient codebook capacity during training. This behavior mirrors the phenomenon observed in language models, where certain "filler" tokens dominate usage in the absence of fine-grained semantic differentiation. Our analysis thus reveals a critical limitation in the codebook’s expressiveness and motivates future work toward more balanced, semantically disentangled discrete representations.

\section{Implications and Insights}
\begin{figure}[tbp]
  \centering
  \begin{subfigure}{0.97\linewidth}
  \includegraphics[width=\linewidth]{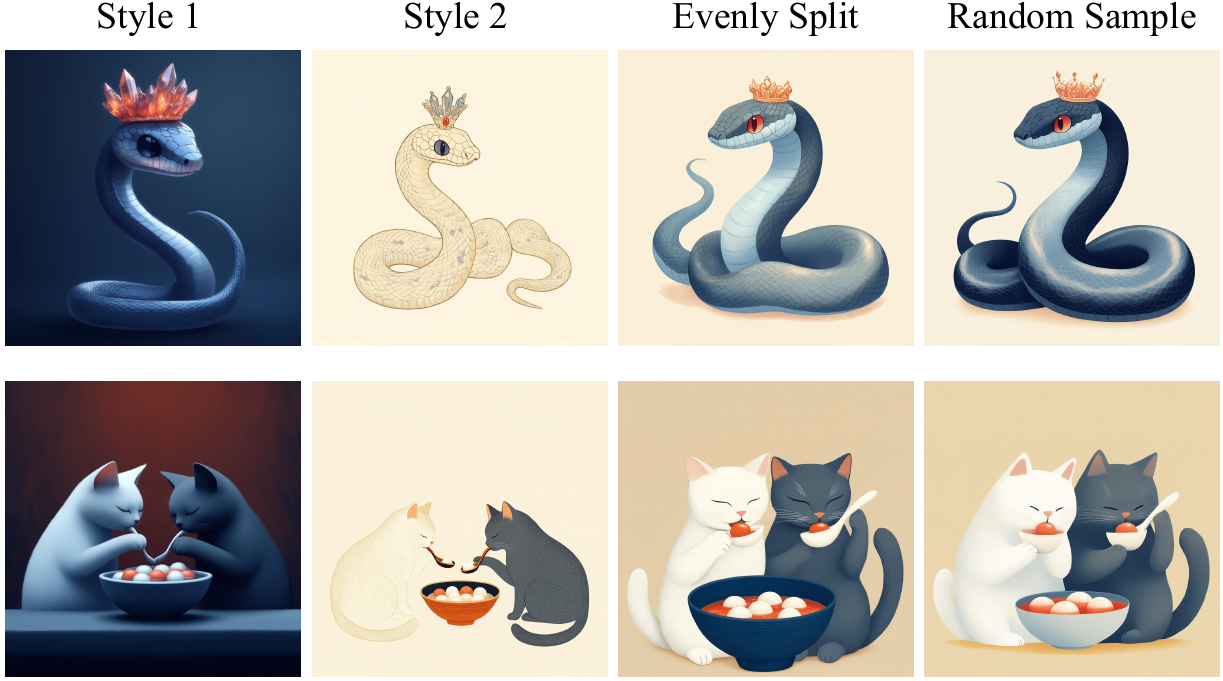}  
  \end{subfigure}
  \caption{
  The impact of token selection strategy on the generation results during style interpolation.}
  \label{fig:tokensample}
\end{figure}

In this paper, we not only propose a novel approach to code-to-style generation but also offer several insightful observations worth further discussion. 

First, we introduce a discrete style codebook to extract and represent stylistic information from images in a discrete latent space. Building upon this representation, we demonstrate that smooth style interpolation can be achieved by blending different style codes. As shown in Fig.~\ref{fig:tokensample}, we observe that the specific token selection strategy during interpolation has minimal impact on the generated results, regardless of whether it is achieved through random sampling or splitting tokens evenly between the two styles (e.g., first half from style A and second half from style B).
This indirectly highlights a key distinction between our learned style representation and traditional image representations: the stylistic information is invariant to the order of its tokens. This property aligns naturally with human intuition about style, which is typically perceived as a holistic attribute rather than a sequence-dependent structure.
Style interpolation is just one emergent application enabled by our learned style representation; the structural properties of this discrete space may facilitate the development of additional, more sophisticated, and practically valuable functionalities. Furthermore, this discrete feature extraction paradigm holds promise for extension to other modalities, such as audio.

Second, we advocate for injecting style information from the textual modality rather than the visual. We argue that human perception of style is inherently semantic rather than purely chromatic; thus, conditioning on text enables the model to capture stylistic attributes that better align with human intuition.

Finally, code-to-style image generation is a task of substantial practical potential, yet it has remained unexplored in the academic literature. We present the first comprehensive framework to address this problem and publicly release both the model weights and source code, which we hope will inspire further research in this emerging direction.

\section{Limitations}
The CoTyle framework represents a significant advancement in the domain of code-to-style image generation. However, several limitations should be acknowledged and explored in future research.

Firstly, while the experimental results indicate that the stylistic diversity of images generated by CoTyle is commendable, it remains somewhat lower than that observed in Midjourney. This outcome is likely linked to the training regimen of the autoregressive style generator, which relies on a discrete style codebook. This codebook, constructed from a limited dataset of curated style-paired images, may not fully encapsulate the extensive variability inherent in human artistic expressions. Expanding the diversity of training data with a broader range of abstract and diverse artistic sources, or implementing more advanced generative architectures like mixture-of-experts models, may enhance the representation of complex, multi-modal style distributions.

Secondly, the resolution and creativity of the generated styles are largely dictated by the quality of the style codebook. Despite utilizing contrastive loss to differentiate styles, the necessity to quantize these into a finite codebook inevitably entails some loss of detailed stylistic subtleties. As a result, certain intricate or hybrid nuances found in reference images may be attenuated or omitted, potentially affecting the expressiveness of the generated outputs.

Finally, although the numerical style code offers a robust framework for reproducibility, the exploration of aesthetically or artistically engaging styles presently involves a stochastic search across a vast code space. Introducing more intuitive mechanisms for navigating or modifying this latent style space could significantly enhance both user agency and practical applicability.

Although CoTyle still has some areas for improvement, as the first open-source work on the task of code-to-style generation, it introduces an overall framework for addressing this task, holding the potential to inspire subsequent research in the field of code-to-style generation.

\section{More Results}
\subsection{Style Interpolation}
\begin{figure*}[tbp]
  \centering
  \begin{subfigure}{0.95\linewidth}
  \includegraphics[width=\linewidth]{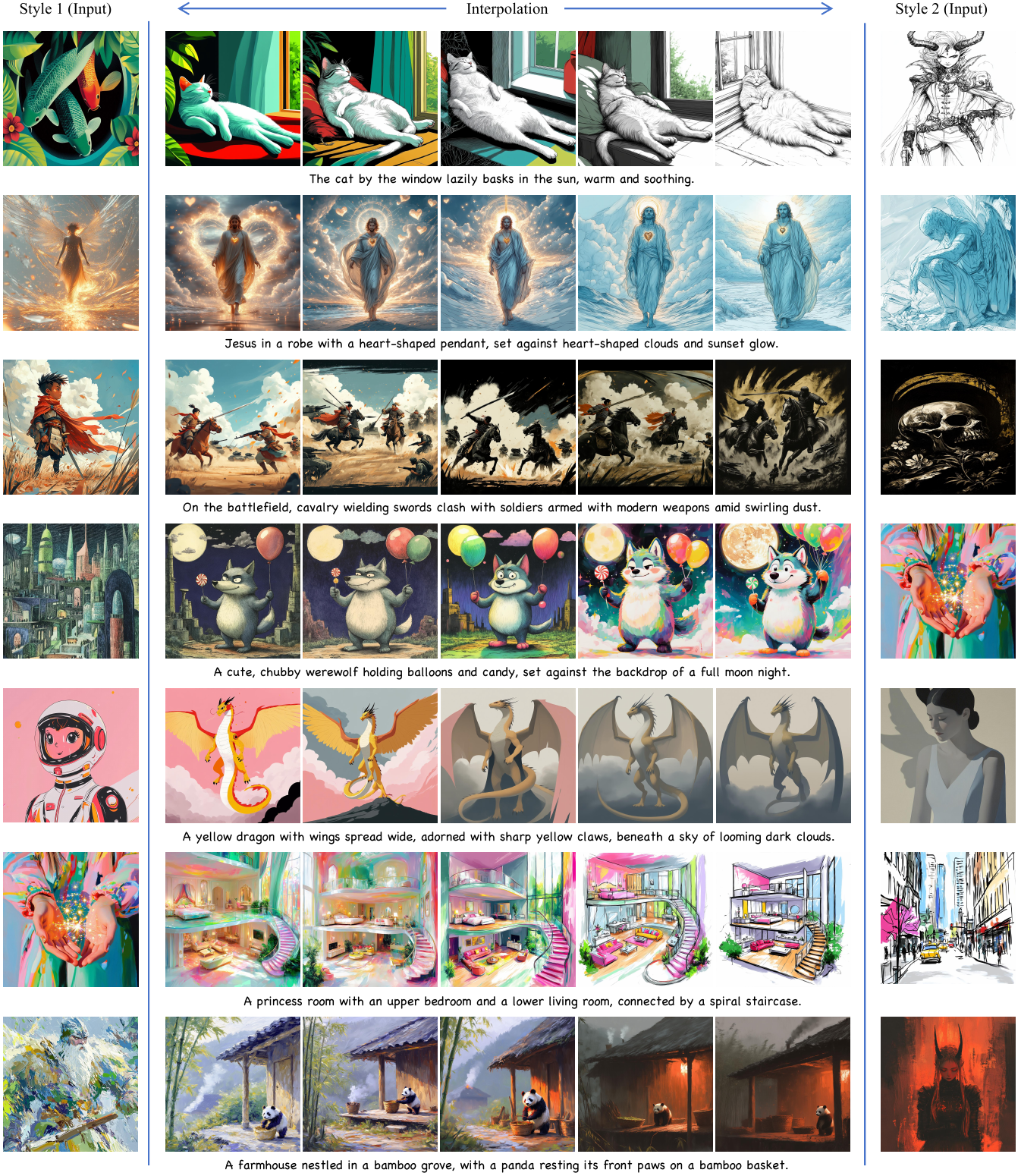}    
  \end{subfigure}
  \caption{More visual results of style interpolation.
  }
  \label{fig:inter}
\end{figure*}

We provide additional style interpolation results in Fig.~\ref{fig:inter}. 
Given two input images, we encode them into two sets of indices using the style codebook and achieve style interpolation by randomly blending the two sets of indices according to a specified mixing ratio.
In Fig.~\ref{fig:inter}, the leftmost and rightmost images represent the source style references, respectively, while the intermediate images show the results of style interpolation generated with a given prompt, using progressively varying blending ratios between the two styles.

\subsection{Code-to-Style Generation}
\begin{figure*}[tbp]
  \centering
  \begin{subfigure}{0.88\linewidth}
  \includegraphics[width=\linewidth]{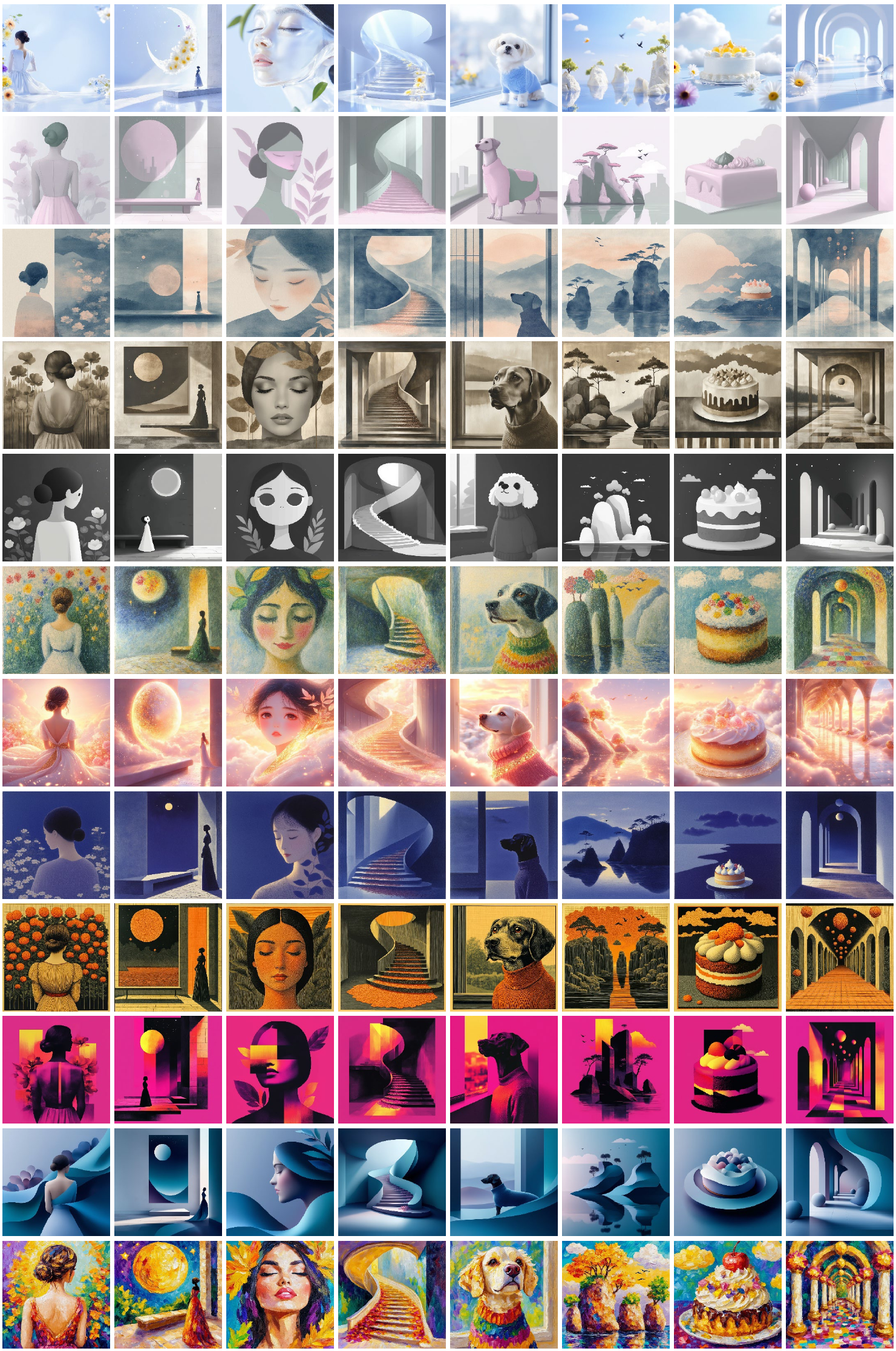}    
  \end{subfigure}
  \caption{More visual results of CoTyle.
  }
  \label{fig:appd1}
\end{figure*}

\begin{figure*}[tbp]
  \centering
  \begin{subfigure}{0.88\linewidth}
  \includegraphics[width=\linewidth]{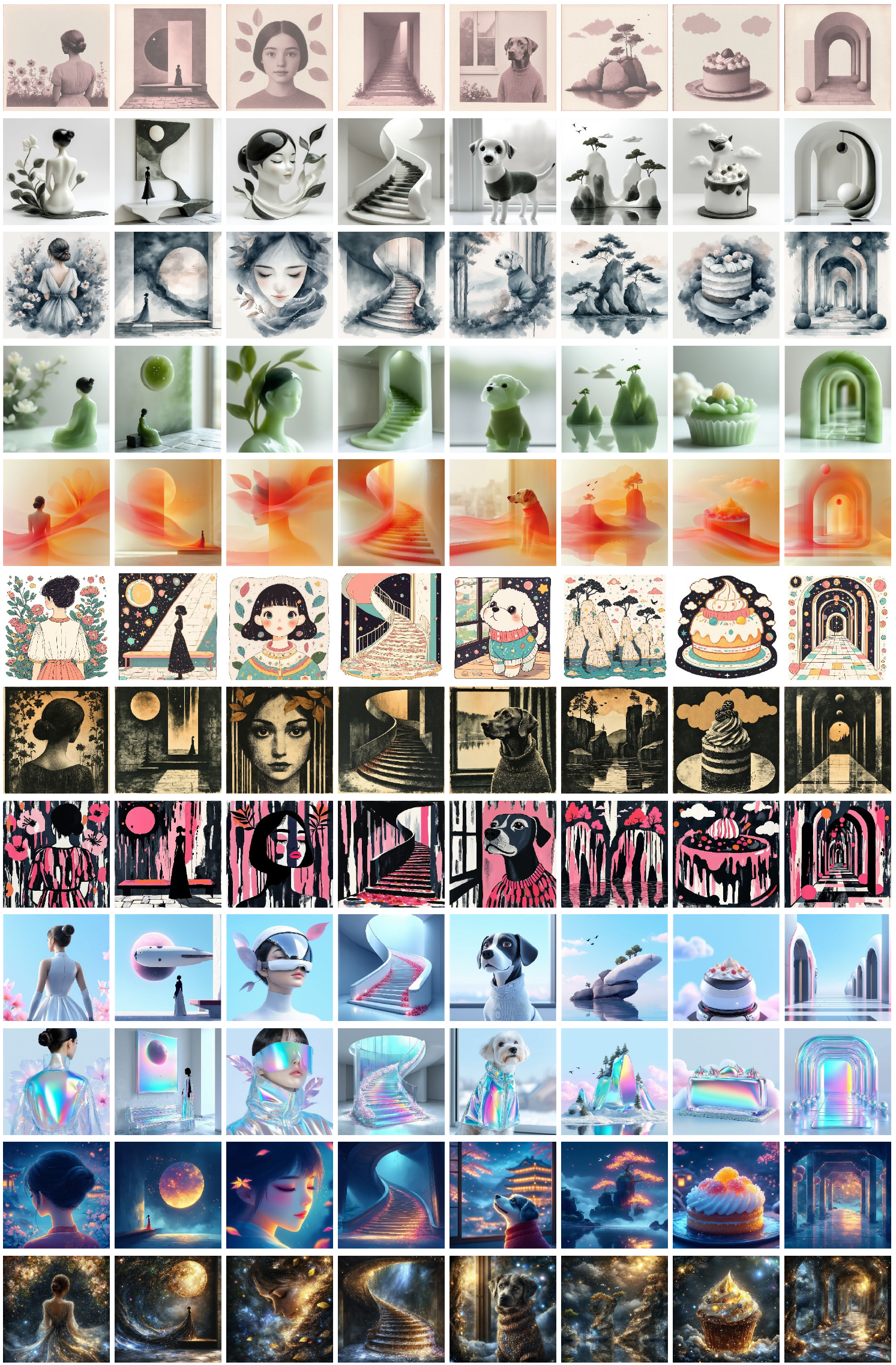}    
  \end{subfigure}
  \caption{More visual results of CoTyle.
  }
  \label{fig:appd2}
\end{figure*}

\begin{figure*}[tbp]
  \centering
  \begin{subfigure}{0.88\linewidth}
  \includegraphics[width=\linewidth]{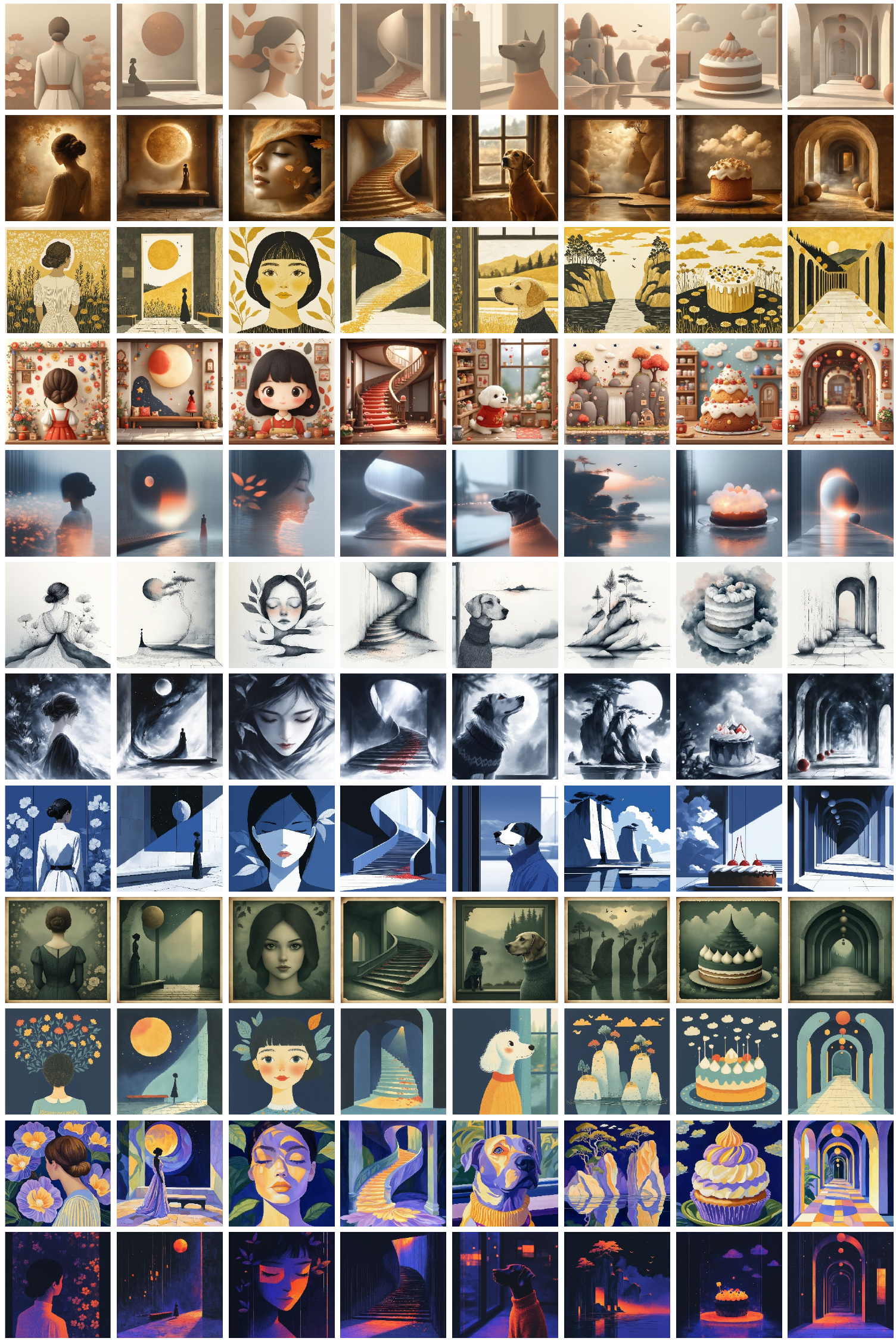}    
  \end{subfigure}
  \caption{More visual results of CoTyle.
  }
  \label{fig:appd3}
\end{figure*}

We present additional code-to-style generation results in Fig.~\ref{fig:appd1}, Fig.~\ref{fig:appd2}, and Fig.~\ref{fig:appd3} to provide a more comprehensive evaluation of our method’s capability in translating structured code inputs into visually coherent stylistic outputs. In each figure, the generation process is carefully controlled: all images within a given row are produced using the same random seed, ensuring that variations across columns arise solely from differences in the input text prompt, rather than stochastic sampling. Conversely, all images within a column are generated under the same textual instruction, allowing direct comparison of stylistic consistency and fidelity across different code conditions. This design enables a clear disentanglement of the influence of code structure versus textual guidance on the final output. The corresponding text prompts for each column are listed in Table~\ref{tab:prompt}. These results further demonstrate the robustness of our approach in mapping diverse code-based styles to rich, semantically aligned visual outputs.

\begin{table}[htbp]
\centering
\caption{Prompts used in visual figures.}
\small
\renewcommand{\arraystretch}{1.6} 
\label{tab:prompts}
\begin{tabular}{%
  >{\centering\arraybackslash}m{0.01\textwidth}%
  >{\raggedright\arraybackslash}m{0.44\textwidth}%
}
\toprule
\multicolumn{1}{c}{\textbf{id}} & \multicolumn{1}{c}{\textbf{prompt}} \\
\midrule
1 & ``A woman in a dress, with her hair tied in a bun, faces away from the viewer. The background is composed of floral patterns.''\\
2 & ``The image shows an artwork featuring a silhouette of a woman in a long dress, standing next to a bench, with an abstract painting in the background containing a large circular object that appears to be the moon.'' \\
3 & ``The image shows a portrait of a woman, with her face obscured by leaves and a few leaves decorating the background.'' \\
4 & ``The image shows a spiral staircase with light streaming down from the top, illuminating the staircase's surface. The staircase is located inside a large building, with the walls and ceiling in the background.'' \\
5 & ``The image shows a dog wearing a sweater, standing by a window, with the scenery outside the window blurred in the background.'' \\
6 & ``The image shows a natural landscape with several uniquely shaped rocks and trees growing in the foreground. In the background is a sky dotted with clouds and two flying birds. The rocks and trees are clearly reflected in the water.'' \\
7 & ``The image shows a cake covered in buttercream and candies, with clouds in the background.'' \\
8 & ``The image shows a corridor made of arches and spheres, with a floor made of blocks and a background.'' \\
\bottomrule
\end{tabular}
\label{tab:prompt} 
\end{table}


\clearpage

\end{document}